\pdfoutput=1

\documentclass[11pt]{article}

\usepackage{EMNLP2023}

\usepackage{times}
\usepackage{latexsym}
\usepackage{tabularx}
\usepackage{booktabs}
\usepackage{tabularx}
\usepackage{rotating} 
\usepackage{geometry}
\usepackage{afterpage}
\usepackage{rotating}
\usepackage{tabularx}
\usepackage{multirow}
\usepackage{booktabs}
\usepackage{adjustbox} 
\usepackage{lipsum} 
\usepackage{rotating}
\usepackage{amsmath}

\usepackage[T1]{fontenc}

\usepackage[utf8]{inputenc}

\usepackage{microtype}

\usepackage{inconsolata}

\usepackage{graphicx} 
\title{Multi-EuP: The Multilingual European Parliament Dataset for Analysis of Bias in Information Retrieval} 

\author{%
  Jinrui Yang\textsuperscript{*} \qquad  Timothy
  Baldwin\textsuperscript{*}\textsuperscript{\dag} \qquad
  Trevor Cohn\textsuperscript{*} \\
  \textsuperscript{*}School of Computing \& Information Systems, The University of Melbourne \\
  \textsuperscript{\dag}Mohamed bin Zayed University of Artificial Intelligence, UAE \\
  {\url{jinruiy@student.unimelb.edu.au}} \qquad  \url{{tbaldwin,trevor.cohn}@unimelb.edu.au}
}

\begin{document}
\maketitle
\begin{abstract}

We present Multi-EuP, a new multilingual benchmark dataset,  comprising 22K multilingual documents collected from the European Parliament, spanning 24 languages. This dataset is designed to investigate fairness in a multilingual information retrieval (IR) context to analyze both language and demographic bias in a ranking context. It boasts an authentic multilingual corpus, featuring topics translated into all 24 languages, as well as cross-lingual relevance judgments. 
Furthermore, it offers rich demographic information associated with its documents, facilitating the study of demographic bias. We report the effectiveness of Multi-EuP for benchmarking both monolingual and multilingual IR. We also conduct a preliminary experiment on language bias caused by the choice of tokenization strategy.

\end{abstract}

\section{Introduction}

Information retrieval (IR) classically uses a retrieval model to query a document collection and return a ranked list of documents which are predicted to be (decreasingly) relevant to the query. Retrieval models have increasingly been based on supervised learning, involving the annotation of documents with relevance scores relative to a given query, and the training of models to predict the relative association between a query and document \citep{karpukhin-etal-2020-dense,DBLP:journals/corr/abs-2004-12832}. 

In parallel with these advances, the democratisation of the internet has led to a surge of individual contributors serving as information disseminators, hailing from various countries and regions, and posting in different languages. This has created possibilities for exploration of cross-lingual and multilingual text retrieval. Cross-lingual retrieval pertains to scenarios where queries are formulated in one language but documents are retrieved from another language. On the other hand, multilingual retrieval involves a query in one language but retrieval of documents across multiple languages simultaneously. An important consideration in any such work is both robustness and fairness across different combinations of languages -- for instance, are results from one language consistently ranked higher than another for certain types of query.

While progress towards multilingual retrieval through the release of datasets such as Mr.\ TYDI \citep{zhang2021mr} and mMARCO \citep{DBLP:journals/corr/abs-2108-13897}, both are limited in that they evaluate monolingual retrieval for a range of languages, rather than true multilingual retrieval, using multiple languages simultaneously. Additionally, mMARCO was created by machine translation of MS MARCO \citep{DBLP:journals/corr/NguyenRSGTMD16}, introducing a confounding factor of  translation errors.

We present a multilingual dataset based on the European Parliament debate archive with queries in 24 distinct languages, and relevance judgements also across all 24 languages. This ensures the ``multilingual'' nature of the dataset in terms of both query-to-document and document-to-query associations. We additionally augment each document with comprehensive metadata of the author, including gender, nationality, political affiliation, and age, for use in exploring fairness with respect to protected attributes.

Our work contributes to the field in three main ways: (1) we construct and release the Multi-EuP dataset, a resource for multilingual retrieval over 24 languages, effectively capturing the multilingual nature of both queries and documents; (2) we explore language bias within the realm of multilingual retrieval, revealing that multilingual IR using BM25 indeed exhibits notable language bias; and (3) we supplement the dataset with rich author metadata to enable research on fairness and demographic bias in IR.\footnote{The Multi-EuP dataset is available for download from \url{https://github.com/jrnlp/Multi-EuP}.}

\section{Background and Related Work}

The European Parliament (EP) serves as an important forum for political debates and decision-making at the European Union level. Members of the European Parliament (MEP) are elected in direct elections across the EU. The European Parliament debate is presided over by the President, who guides MEPs in discussing specific subjects. 

EP debates have been the source of three key datasets. First, \textit{Europarl-2005} was crafted by \citet{koehn-2005-europarl} by collecting EP debates documents from 1996 to 2011, and extracting translations as a parallel corpus for statistical machine translation,  enriched with attributes including \textit{debate date, chapter id, MEP id, language, MEP name}, and \textit{MEP party}. 


Later,  \citet{rabinovich-etal-2017-personalized} built \textit{Europarl-2017} upon \textit{Europarl-2005}, by introducing additional demographic attributes: \textit{MEP gender} and \textit{MEP age}. These were sourced from sources such as Wikidata \citep{42240} and automatic annotation tools such as \textit{Genderize}\footnote{\url{https://genderize.io/}} and \textit{AlchemyVision}.\footnote{\url{https://www.ibm.com/smarterplanet/us/en/ibmwatson/developercloud/alchemy-vision.html}} However, \textit{Europarl-2017} is limited to only two language pairs: English--German and English--French. \textit{Europarl-2018} \citep{vanmassenhove-hardmeier-2018-europarl} expanded upon \textit{Europarl-2017} to add twenty additional language pairs, based on the manual translations in the EP archives. These corpora have been used primarily for machine translation research. 

Since 2020, the EU has publicly released raw debates in the form of transcribed source-language speeches with rich multilingual topic index data, along with the original video and audio recordings.  This forms the basis of the Multi-EuP dataset, with additional attributes for each speaking MEP such as an image, birthplace, and nationality.

\citet{zhang2021mr} introduced Mr.\ TYDI, an evaluation benchmark dataset for dense retrieval assessment over 11 languages. This dataset is constructed from  TYDI \citep{clark-etal-2020-tydi}, a question answering dataset. For each language, annotators assign relevance scores as judgments for questions, derived from Wikipedia articles. Notably, the questions for different languages are crafted independently, and relevance judgements are provided in-language only. Based on the dataset, the authors evaluate on monolingual retrieval tasks for non-English languages using BM25 and mDPR as zero-shot baselines. However, Mr.\ TYDI's scope is limited in that it is not truly multilingual, in that queries in a given language are only performed over documents in that language. This is part of the void our work aims to address.

MS MARCO \citep{DBLP:journals/corr/NguyenRSGTMD16} is a widely-used dataset, sourced from Bing's search query logs, but for English queries and documents only. To mitigate this, \citet{DBLP:journals/corr/abs-2108-13897} introduced mMARCO, a multilingual variant of the MS MARCO passage ranking dataset, spanning 13 languages and created through machine translation, based on one open-source approach \citep{tiedemann-thottingal-2020-opus} and one commercial system in the form of Google Translate.\footnote{\url{https://cloud.google.com/translate}} Analysis of the authors' results reveals a positive correlation between translation quality and retrieval performance, with higher translation BLEU scores yielding improved retrieval MRR outcomes. However, similar to Mr.\ TYDI, mMARCO focuses on in-language retrieval only for multiple languages, rather than multilingual retrieval.

Throughout the past few decades, numerous datasets and tasks pertaining to multilingual retrieval have been developed for evaluation, through efforts such as CLEF, TREC, and FIRE, each contributing standardized document collections and evaluation procedures. These evaluation datasets facilitate genuine multilingual IR research such as \citet{article} and \citet{lawrie2023neural}. However, the scope of these datasets is generally limited to a small number of queries. For example, in the case of CLEF 2001-2003, each edition encompasses a mere few dozen queries. This limitation tends to confine research predominantly to evaluation and not offer a resource for training a multilingual ranking model. Our dataset is of a scale to accommodate both large-scale training and evaluation of multilingual retrieval methods.

Compared with the related work above, our work augments the multilingual mixture of queries and documents compared to Mr.TYDI, preserves the authenticity of multilingual contexts compared to mMARCO's translation-based approach, and surpasses the query count limitations of tasks like CLEF.

\section{Multi-EuP}

In our approach, we consider the debate topics to be the queries, and the text of each individual speech delivered by an MEP to be a document.


\begin{table*}[t]
\centering
\renewcommand{\arraystretch}{0.95}
\begin{tabularx}{\linewidth}{Xl}
\begin{tabular}{@{} m{1.7cm}m{0.9cm}m{5.1cm}m{1cm}m{1cm}m{1.5cm}m{1.5cm} @{}}
\toprule
\textbf{Language} & \textbf{ISO code} & \textbf{Countries where official lang.} & \textbf{Native Usage} & \textbf{Total Usage} & \textbf{\# Docs} & \textbf{Words per Doc} \\
\midrule
English & EN & United Kingdom, Ireland, Malta & 13\% & 51\% & 7123 & 286/200 \\
German & DE & Germany, Belgium, Luxembourg & 16\% & 32\% & 3433 & 180/164 \\
French & FR & France, Belgium, Luxembourg & 12\% & 26\% & 2779 & 296/223 \\
Italian & IT & Italy  & 13\% & 16\% & 1829 & 190/175 \\
Spanish & ES & Spain  & 8\% & 15\% & 2371 & 232/198 \\
Polish & PL & Poland & 8\% & 9\% & 1841 & 155/148 \\
Romanian & RO & Romania  & 5\% & 5\% & 794 & 186/172 \\
Dutch & NL & Netherlands, Belgium & 4\% & 5\% & 897 & 184/170 \\
Greek & EL & Greece, Cyprus & 3\% & 4\% & 707 & 209/205 \\
Hungarian & HU & Hungary & 3\% & 3\% & 614 & 126/128 \\
Portuguese & PT & Portugal & 2\% & 3\% & 1176 & 179/167 \\
Czech & CS & Czech Republic & 2\% & 3\% & 397 & 167/149 \\
Swedish & SV & Sweden & 2\% & 3\% & 531 & 175/165 \\
Bulgarian & BG & Bulgaria & 2\% & 2\% & 408 & 196/178 \\
Danish & DA & Denmark & 1\% & 1\% & 292 & 218/198 \\
Finnish & FI & Finland & 1\% & 1\% & 405 & 94/87 \\
Slovak & SK & Slovakia & 1\% & 1\% & 348 & 151/158 \\
Lithuanian & LT & Lithuania & 1\% & 1\% & 115 & 142/127 \\
Croatian & HR & Croatia & <1\% & <1\% & 524 & 183/164 \\
Slovene & SL & Slovenia & <1\% & <1\% & 270 & 201/163 \\
Estonian & ET & Estonia & <1\% & <1\% & 58 & 160/158 \\
Latvian & LV & Latvia & <1\% & <1\% & 89 & 111/123 \\
Maltese & MT & Malta & <1\% & <1\% & 178 & 117/115 \\
Irish & GA & Ireland & <1\% & <1\% & 33 & 198/172 \\
\bottomrule
\end{tabular}
\caption{Multi-EuP statistics, broken down by language: ISO language code; EU member states using the language officially; proportion of the EU population speaking the language \citep{chalkidis-etal-2021-multieurlex}; number of debate speech documents; and words per document (mean/median).}
\label{Multi-EuP-stats}
\end{tabularx}
\end{table*}

\paragraph{Topics}

The topics are officially annotated by the EU, and professionally translated into 24 different languages.\footnote{\url{https://www.europarl.europa.eu/translation/en/translation-at-the-european-parliament/}} During preprocessing, we filter out procedural debate topics such as \textit{agenda}, leaving 1.1K unique topics. They will serve as a valuable resource for assessing language bias in multilingual ranking methods, given that all the topics across different languages are semantically consistent. 

\paragraph{Documents}
The 22K multilingual documents within the Multi-EuP dataset originate from MEP speeches during parliamentary debates.  Each document annotated with additional metadata,  including the date of the speech, the MEP ID, and a link to the video recording for potential multimodal research but not used here. Table \ref{Multi-EuP-stats} shows a detailed breakdown of the language distribution and descriptive statistics of the dataset. We include in our corpus documents only in the original language, as spoken by the MEP, but not their translations into other languages. Our only use of translations is the debate topics themselves.

\paragraph{Judgments} To assess the relevance of documents to a given query, we use a binary relevance judgment, based on whether the speech was part of a debate on the given topic, resulting in one positive relevance judgment per document, meaning that the document collection is much less sparse than Mr.\ TYDI and MS MARCO, for example.

\paragraph{Languages}
Multi-EuP covers 24 EU languages from seven families (Germanic, Romance, Slavic, Uralic,
Baltic, Semitic, Hellenic), each of which is the official language of one or more member states. Table \ref{Multi-EuP-stats} provides a breakdown of each language's EU usage, member state distribution, and population, using ISO-639 codes.

\paragraph{MEP}

\begin{figure*}[t]
    \centering
    \includegraphics[width=\textwidth]{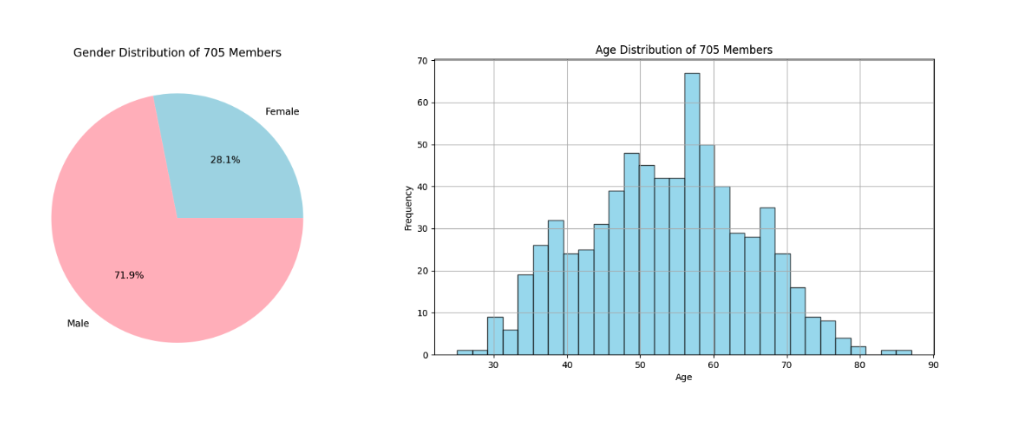}
    \caption{The gender and birth year distributions of the 705 MEPs in Multi-EuP dataset. The birth year corresponds to the current age calculation.}
    \label{fig:gender_age}
\end{figure*}


Multi-EuP encompasses 705 members elected across the 27 member states of the EU. We constructed the MEP dictionary by collecting MEP attributes such as \textit{name, photo, id in EU, nationality, place of birth, party affiliation,} and \textit{spoken language}. We further annotated MEPs with gender and their birthdate, based on Wikipedia profiles and \citet{rabinovich-etal-2017-personalized}, and manually checked if difference existing.  Figure \ref{fig:gender_age} illustrates the gender and age distribution across MEPs, with male MEPs being more than twice as numerous as female MEPs, and the majority falling within the 40--70 age range. This corpus is rare, perhaps unique, due to its richly detailed speaker demographic information, which enables  research on fairness and bias in information retrieval.



\paragraph{Data Split}For data splitting, we select two sets with 100 language-specific and distinct topics for development and test set in 24 languages, and keep the remaining topics to the training set. This design choice was made to maintain an ample supply of topics and judgment samples essential for the training of deep learning models, and also facilitate subsequent cross-lingual comparative research.

\paragraph{Supported Task}

Similarly to Mr.TYDI \citep{zhang2021mr}, Multi-EuP can be used for monolingual retrieval in English as well as non-English languages (eg. Swedish queries against Swedish documents). However, unlike Mr.TYDI, Multi-EuP encompasses multilingual documents and identical multilingual topics, ensuring that queries in different languages can be compared.  Consequently, Multi-EuP can support diverse information retrieval experimental tasks. These including \textit{one-vs-one} scenarios with single one language queries against single one language documents, in other words, monolingual or cross-lingual IR, \textit{one-vs-many} scenarios with single-language queries against multilingual documents, i.e., multilingual IR, and \textit{many-vs-many} scenarios involving multilingual queries against multilingual documents, i.e, mixed multilingual IR).

\section{Experiments and Findings}

\begin{table*}[t]
\centering
\footnotesize 
\begin{tabularx}{\linewidth}{Xl}
\small 
\begin{tabular}{m{0.7cm} *{20}{c@{\,\,\,}}c }
\toprule
\textbf & \multicolumn{5}{c}{\textbf{GERMANIC}} &&  \multicolumn{5}{c}{\textbf{ROMANCE}} && \multicolumn{3}{c}{\textbf{SLAVIC}} && \multicolumn{2}{c}{\textbf{URALIC}}  &  \\
\cmidrule{2-6}
\cmidrule{8-12}
\cmidrule{14-16}
\cmidrule{18-19}
\textbf & \textbf{EN} & \textbf{DA} & \textbf{DE} & \textbf{NL} & \textbf{SV} && \textbf{RO} & \textbf{ES} & \textbf{FR} & \textbf{IT} & \textbf{PT} && \textbf{PL} & \textbf{BG} & \textbf{CS} && \textbf{HU} & \textbf{FI} && \textbf{EL} \\
\midrule
\multicolumn{21}{l}{\textbf{One-vs-one} (Queries and documents in the same language.)} \\
\midrule

num$_q$ &  839 &  208 &  840 & 458 & 330 && 434 &  680 &  765 &  659 &  557 && 628 & 273 & 259  && 404 & 251 && 360 \\
num$_d$ & 7123 & 268 & 3433 & 897 & 531 && 794 & 2371 & 2779 & 1829 & 1176 && 1841 & 408 & 397 && 614 & 405 && 707 \\
num$_r$ & 7123 & 3433 & 3433 & 897 & 531 && 794 & 2371 & 2779 & 1829 & 1176 && 1841 & 408 & 397 && 614 & 405 && 707 \\
\textbf{MRR} & \textbf{73.79} & \textbf{45.51} & \textbf{56.70} & \textbf{39.02} & \textbf{45.77} && \textbf{42.58} & \textbf{54.25} & \textbf{46.57} & \textbf{56.40} & \textbf{56.51} && \textbf{47.68} & \textbf{44.70} & \textbf{47.12} && \textbf{39.99} & \textbf{46.46} && \textbf{39.58} \\
\midrule
\multicolumn{21}{l}{\textbf{One-vs-many} (A fixed set of queries in the given language, with documents in all 24 languages.)} \\
\midrule

num$_q$ & 100 & 100 & 100 & 100 & 100 && 100 & 100 & 100 & 100 & 100 && 100 & 100 & 100 && 100 & 100 && 100 \\
num$_d$ & 22K & 22K & 22K & 22K & 22K && 22K & 22K & 22K & 22K & 22K && 22K & 22K & 22K && 22K & 22K && 22K \\
num$_r$ &2902 & 1997 & 1995 & 1992 & 1996 && 1996 & 1996 & 1924 & 1994 & 1997 && 1996 & 1996 & 1997 && 1996 & 1997 && 1996 \\
\textbf{MRR} & \textbf{62.79} & \textbf{16.15} & \textbf{28.27} & \textbf{20.88} & \textbf{19.40} && \textbf{16.10} & \textbf{22.57} & \textbf{24.22} & \textbf{14.24} & \textbf{18.7} && \textbf{4.80} & \textbf{7.57} & \textbf{9.52} && \textbf{7.51} & \textbf{17.61} && \textbf{11.16} \\

\bottomrule
\end{tabular}
\caption{Details of Multi-EuP for the 16 most widely spoken EU official languages, in terms of the number of queries ($q$), documents ($d$) and relevance judgements ($r$). Results are for BM25 in one-vs-one and one-vs-many settings based on MRR$@$100 (\%). See Table \ref{AP:Multi-EuP-stats} in the Appendix for results across all languages. Note that as each document has a unique topic which in turn defines the relevance judgements, num$_d$= num$_r$ in the one-vs-one setting.}
\label{Multi-EuP-BM25}
\end{tabularx}
\end{table*}



We conduct preliminary experiments in both one-vs-one and one-vs-many settings, as described above.

\paragraph{Methods} We base our experiments on BM25 with default settings ($k_1$ = 0.9 and $b$ = 0.4), a popular traditional information retrieval baseline. Our implementation is based on Pyserini \citep{inproceedingslin}, which is built upon Lucene \citep{yang+:2017}. Notably, the latest LUCENE 8.5.1 API offers language-specific tokenizers, \footnote{Provided by the Anaylzer package in LUCENE. \url{https://lucene.apache.org/core/8_5_1/analyzers-common/index.html}} covering 19 out of the 24 languages present in Multi-EuP. For the remaining languages --- namely Polish (PL), Croatian (HR), Slovak (SK), Slovenian (SL), and Maltese (MT) --- we use a whitespace tokenizer. 


\paragraph{Evaluation}
Our primary evaluation metric is Mean Reciprocal Rank (MRR). For a single query, the reciprocal rank is 
$RR = \frac{1}{\text{rank}}$ where ${\text{rank}}$ is the position of the highest-ranked relevant document. If no correct answer was returned, then the reciprocal rank is defined to be 0. For multiple queries $Q$, the MRR is the mean of the $Q$ reciprocal ranks.\[
MRR = \frac{1}{Q} \sum_{i=1}^Q \frac{1}{\text{rank}_i} \]
MRR@$k$ denotes MRR computed at a depth of $k$ results. Note that the higher the number the better, and that a perfect retriever achieves an MRR of 1 (assuming every query has at least one relevant document). The choice of setting $k = 100$ aligns with prior endeavors over MS MARCO \citep{DBLP:journals/corr/NguyenRSGTMD16}.


\subsection{Monolingual IR (\textit{one-vs-one)}}
\paragraph{Experimental Setup} We first present results over Multi-EuP in a monolingual setting across the 24 different languages. Specifically, we evaluate single-language queries against documents in the same language. In this configuration, we partitioned our original collection of 22K documents into 24 distinct language-specific sub-collections. Table \ref{Multi-EuP-BM25} presents the results broken down across languages. 

\paragraph{Results and Findings} Table \ref{Multi-EuP-BM25} presents the MRR@100 results for BM25 on Multi-EuP. There are two high-level findings:

First, Multi-EuP is a relatively easy benchmark for monolingual information retrieval, as the MRR@100 is always around 40 or greater (meaning that the first relevant document is in the top-3 results on average). Indeed, the average MRR across the 24 test languages is 49.61. While direct comparison is not possible, it is noteworthy that for Mr.\ TYDI, the average MRR is 32.1 across 11 languages. Part of this difference can be attributed to the fact that our relevance judgments are not as sparse as theirs. 

Second, similar to Mr.\ TYDI, direct comparison of absolute scores between languages is not meaningful in a monolingual setting, as the document collection size differs. 




\subsection{Multilingual IR (\textit{one-vs-many)}}
\paragraph{Experimental Setup} In contrast to Mr.\ TYDI \citep{zhang2021mr}, Multi-EuP supports one-vs-many retrieval, and allows us to systematically explore the effect of querying the same document collection with the same set of topics in different languages. This is because we have translations of the topics in all languages, documents span multiple languages, and judgments are cross-lingual (e.g., English queries potentially yield relevant Polish documents). For this experiment, we use the default whitespace tokenizer in the Pyserini library.

\paragraph{Results and Findings} 
Table \ref{Multi-EuP-BM25} presents the MRR results for BM25 for multilingual information retrieval on 100 topics from the Multi-EuP test set. It's worth noting that these topics have translation-equivalent content in the different languages. Consequently, the one-vs-many approach allows us to analyze language bias. We made several key observations:

First, unsurprisingly, having more relevance judgments tends to improve ranking accuracy. Therefore, when comparing English topics with other languages, English exhibits notably better MRR performance. 


Second, despite there being consistency in the topics, document collection, and relevance judgments, there is a significant disparity in MRR scores across languages, an effect we investigate further in the next section.




\begin{figure*}[t]
    \centering
    \includegraphics[width=1\textwidth]{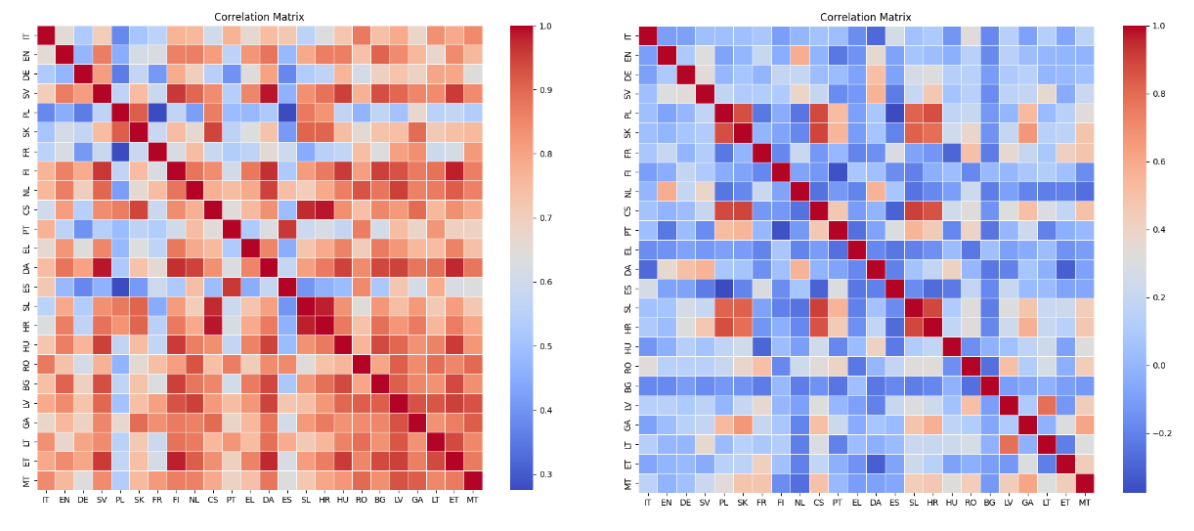}
    \caption{Language correlation matrix between topics and the ranking output top 100 relevant documents in a one-vs-many setting. The row is the topic languages, the columns is the document languages. The left matrix displays results using a language-specific tokenizer, while the right matrix represents the experiment with a simple whitespace tokenizer. Both of them show strong language bias between the language of the topic and the retrieved documents. 
}
    \label{fig:correlation}
\end{figure*}

\section{Language Bias Discussion}
In light of our findings in a one-vs-many setting, we were keen to delve further into the underlying causes of the disparity between languages. 

\subsection{Bias Detection}

Language bias is likely if the query language aligns better with one document language than another. As mentioned earlier, Pyserini supports different tokenizers, specifically language-specific tokenizers or simple whitespace tokenization. Therefore, in the one-vs-many 
setting, we analyze the composition of the top-100 rankings for the 100 topics. During indexing of the document collection, we used the simple whitespace tokenizer, given the multilingual nature of the collection. However, over the queries during retrieval, we employed two different tokenizers --- a language-specific tokenizer, and the whitespace tokenizer.



We conducted a correlation analysis between the language of the topics and the language of the top 100 relevant documents. From Table \ref{Multi-EuP-BM25}, we can see that relevance judgments in our test cases are consistent across languages, ensuring uniformity in the correlation matrix within the test set. However, Figure \ref{fig:correlation} reveals that both approaches generate strong language bias. In both cases, the query language aligns better with documents in its own language than others. The right plot appears to show that languages from the same family has strong correlation (e.g., PL, CS) and (IT, ES) since they may have some shared vocabulary.



\begin{figure*}[t]
    \centering
    \includegraphics[width=1\textwidth]{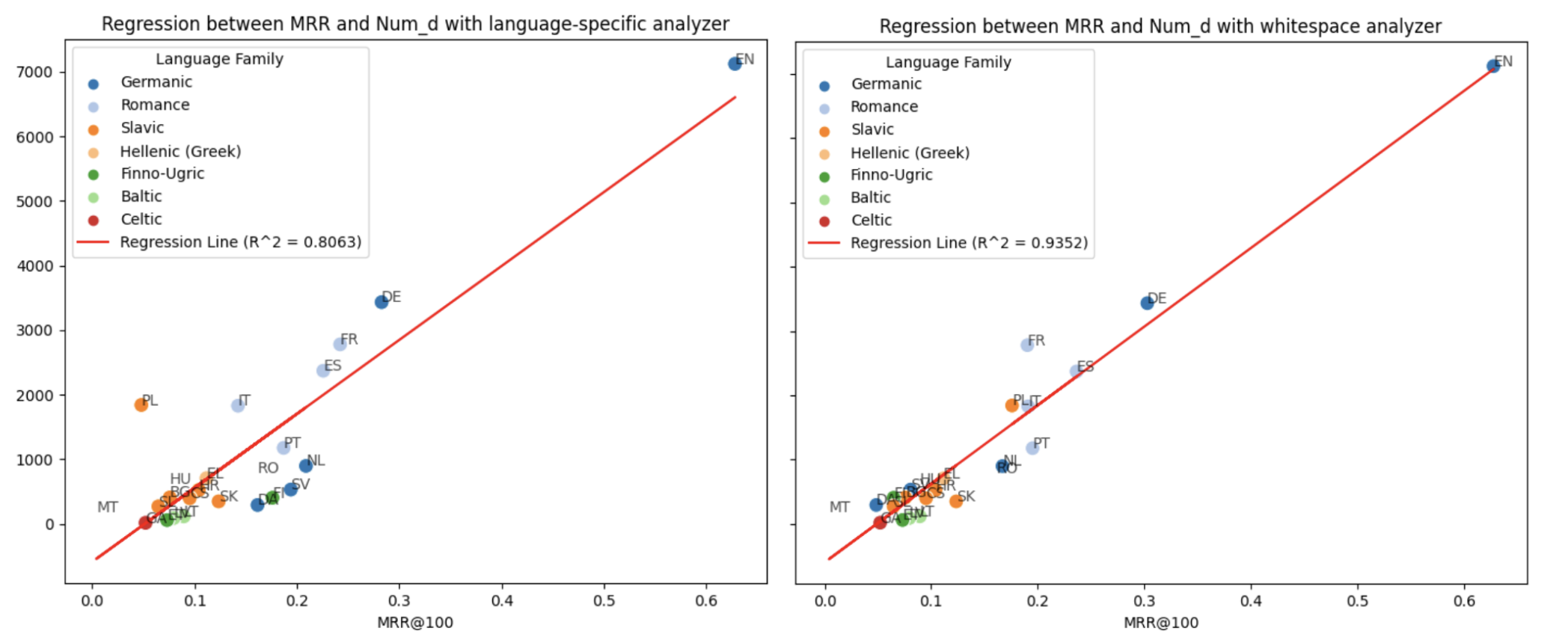}
    \caption{Linear regression between MRR@100 and the number of documents per language. The left plot is based on collection indexing with a whitespace tokenizer but a language-specific tokenizer over the queries. The right plot uses a whitespace tokenizer for both indexing the collection and the queries. The higher $R^2$ for the right plot suggests that using a whitespace tokenizer for both the collection and queries reduces language bias in multilingual IR.}
    \label{fig:mrr}
\end{figure*}

\subsection{Collection Distribution Factors}
Initially, we hypothesized that the disparity for each language may be a contributing factor to this bias. Figure \ref{fig:mrr} presents the regression line between the number of documents in a given language and MRR, which explains much of the variation across languages. 

However, note the outlier above the regression line (Polish: PL), which has a substantial number of documents but surprisingly low MRR performance. We refer to this phenomenon as a ``BM25 unfriendly'' language. According to \citet{wojtasik2023beirpl}, the main reason for the low performance of Polish lies in its highly-inflected morphology, giving rise to a a multitude of word forms per lexeme, including inflections of proper names, and complex morphological structure. In such cases, lexical matching is less effective than in other morphologically-simpler languages. Furthermore, LUCENE 8.5.1 API does not have a language-specific tokenizer for Polish.  Conversely, languages below the regression line can be termed ``BM25 friendly'' languages, as they require fewer documents to achieve higher MRR in retrieval.

\subsection{Language Tokenizer Factors}
Secondly, we speculated that the choice of language-specific Analyzer in LUCENE might be a contributing factor, as it influences word tokenization, token filter, synonym expansion and other processing. \footnote{\url{https://lucene.apache.org/core/8_0_0/core/org/apache/lucene/analysis/package-summary.html\#package.description}} To investigate this, we conducted a controlled experiment in the one-vs-many setting. When indexing the collection, given the multilingual nature of the collection, we employed whitespace as the tokenizer. However, over the queries, we experimented with either a language-specific tokenizer or whitespace tokenizer. We then compared the linear regression of MRR against the number of documents in Figure \ref{fig:mrr}.  On the right side of the plot, we can see a strong correlation when using whitespace tokenization for both the collection and the queries, reducing language bias.

Furthermore, when transitioning from language-specific tokenizers to whitespace tokenizers, the overall MRR across all languages declined modestly, from 15.02 to 14.18. That is, the original performance level was largely preserved, but language bias was diminished in using simple whitespace tokenization.



\section{Conclusion}
In this paper, we introduce Multi-EuP, a novel dataset for multilingual information retrieval across 24 languages, collected from European Parliament debates. The demographic information provided by the Multi-EuP dataset serves a dual purpose: not only does it contribute to multilingual retrieval tasks, but it also holds significant potential for advancing research in the realm of fairness and bias. This dataset can play a pivotal role in investigating issues of equitable representations and mitigation of biases within document ranking settings.

Multi-EuP facilitates diverse information retrieval (IR) scenarios, encompassing one-vs-one, one-vs-many, and many-vs-many settings. We demonstrated the utility of Multi-EuP as a benchmark for evaluating both monolingual and multilingual IR. Our study reveals the presence of language bias in multilingual IR when employing BM25. We further validate the effectiveness of mitigating this bias through the strategic implementation of whitespace as a language tokenizer.

We propose to conduct future work in three main areas. First, we intend to expand our investigation of language bias to encompass a broader range of ranking methods, including neural methods such as mDPR \cite{zhang2021mr}, mColBERT \citep{lawrie2023neural} and PLAID-X\citep{10.1145/3511808.3557325}. Second, we will expand the dataset by developing an automated API to retrieve data published by the European Parliament (EP), thereby ensuring real-time synchronization of our dataset. Lastly, our current experiments have explored language bias only, but we plan to further investigate gender bias, age bias, and nationality bias.


\section*{Limitations}
The limitations of the Multi-EuP dataset are notable but navigable. Primarily, the temporal coverage of the dataset is confined to the past three years. This temporal constraint arises due to the fact that, preceding 2020, documents released by the EU were predominantly available in mono-lingual versions only. However, a potential remedy lies in the amalgamation of the Europarl \citep{koehn-2005-europarl} collection, enabling a more comprehensive and holistic Multi-EuP dataset.

Furthermore, it is worth noting the domain skew of the dataset, in that Multi-EuP inevitably centers on political matters. While this presents challenges, particularly in terms of the intricate nuances of political language, it inherently serves as an excellent foundational stepping stone for delving into the intricacies of multilingual retrieval. We believe, however, that this dataset can serve as a launching pad for broader explorations encompassing cross-domain and open-domain transfer learning scenarios, thus contributing to the broader landscape of language understanding and retrieval.

\section*{Ethics Statement}
The dataset contains publicly-available EP data
that does not include personal or sensitive information, with the exception of information
relating to public officeholders, e.g., the names of the active members of the European Parliament, European Council, or other official administration
bodies. The collected data is licensed under the
Creative Commons Attribution 4.0 International
licence. \footnote{\url{https://eur-lex.europa.eu/content/legal-notice/legal-notice.html}}

\section*{Acknowledgements}
This research was funded by Melbourne Research Scholarship and undertaken using the LIEF HPC-GPGPU Facility hosted at the University of Melbourne. This facility was established with the assistance of LIEF Grant LE170100200. We would like to thank George Buchanan for providing valuable feedback.

\bibliography{emnlp2023}

\begin{thebibliography}{18}
\expandafter\ifx\csname natexlab\endcsname\relax\def\natexlab#1{#1}\fi

\bibitem[{Bonifacio et~al.(2021)Bonifacio, Campiotti, de~Alencar~Lotufo, and Nogueira}]{DBLP:journals/corr/abs-2108-13897}
Luiz~Henrique Bonifacio, Israel Campiotti, Roberto de~Alencar~Lotufo, and Rodrigo~Frassetto Nogueira. 2021.
\newblock \href {http://arxiv.org/abs/2108.13897} {{mMARCO}: {A} multilingual version of {MS} {MARCO} passage ranking dataset}.
\newblock \emph{CoRR}, abs/2108.13897.

\bibitem[{Chalkidis et~al.(2021)Chalkidis, Fergadiotis, and Androutsopoulos}]{chalkidis-etal-2021-multieurlex}
Ilias Chalkidis, Manos Fergadiotis, and Ion Androutsopoulos. 2021.
\newblock \href {https://doi.org/10.18653/v1/2021.emnlp-main.559} {{M}ulti{EURLEX} - a multi-lingual and multi-label legal document classification dataset for zero-shot cross-lingual transfer}.
\newblock In \emph{Proceedings of the 2021 Conference on Empirical Methods in Natural Language Processing}, pages 6974--6996, Online and Punta Cana, Dominican Republic. Association for Computational Linguistics.

\bibitem[{Clark et~al.(2020)Clark, Choi, Collins, Garrette, Kwiatkowski, Nikolaev, and Palomaki}]{clark-etal-2020-tydi}
Jonathan~H. Clark, Eunsol Choi, Michael Collins, Dan Garrette, Tom Kwiatkowski, Vitaly Nikolaev, and Jennimaria Palomaki. 2020.
\newblock \href {https://doi.org/10.1162/tacl_a_00317} {{T}y{D}i {QA}: A benchmark for information-seeking question answering in typologically diverse languages}.
\newblock \emph{Transactions of the Association for Computational Linguistics}, 8:454--470.

\bibitem[{Karpukhin et~al.(2020)Karpukhin, Oguz, Min, Lewis, Wu, Edunov, Chen, and Yih}]{karpukhin-etal-2020-dense}
Vladimir Karpukhin, Barlas Oguz, Sewon Min, Patrick Lewis, Ledell Wu, Sergey Edunov, Danqi Chen, and Wen-tau Yih. 2020.
\newblock \href {https://doi.org/10.18653/v1/2020.emnlp-main.550} {Dense passage retrieval for open-domain question answering}.
\newblock In \emph{Proceedings of the 2020 Conference on Empirical Methods in Natural Language Processing (EMNLP)}, pages 6769--6781, Online. Association for Computational Linguistics.

\bibitem[{Khattab and Zaharia(2020)}]{DBLP:journals/corr/abs-2004-12832}
Omar Khattab and Matei Zaharia. 2020.
\newblock \href {http://arxiv.org/abs/2004.12832} {Colbert: Efficient and effective passage search via contextualized late interaction over {BERT}}.
\newblock \emph{CoRR}, abs/2004.12832.

\bibitem[{Koehn(2005)}]{koehn-2005-europarl}
Philipp Koehn. 2005.
\newblock \href {https://aclanthology.org/2005.mtsummit-papers.11} {{E}uroparl: A parallel corpus for statistical machine translation}.
\newblock In \emph{Proceedings of Machine Translation Summit X: Papers}, pages 79--86, Phuket, Thailand.

\bibitem[{Lawrie et~al.(2023)Lawrie, Yang, Oard, and Mayfield}]{lawrie2023neural}
Dawn Lawrie, Eugene Yang, Douglas~W. Oard, and James Mayfield. 2023.
\newblock \href {http://arxiv.org/abs/2209.01335} {Neural approaches to multilingual information retrieval}.
\newblock arXiv cs.IR 2209.01335.

\bibitem[{Lin et~al.(2021)Lin, Ma, Lin, Yang, Pradeep, and Nogueira}]{inproceedingslin}
Jimmy Lin, Xueguang Ma, Sheng-Chieh Lin, Jheng-Hong Yang, Ronak Pradeep, and Rodrigo Nogueira. 2021.
\newblock Pyserini: A {Python} toolkit for reproducible information retrieval research with sparse and dense representations.
\newblock \url{https://github.com/castorini/pyserini}.

\bibitem[{Nguyen et~al.(2016)Nguyen, Rosenberg, Song, Gao, Tiwary, Majumder, and Deng}]{DBLP:journals/corr/NguyenRSGTMD16}
Tri Nguyen, Mir Rosenberg, Xia Song, Jianfeng Gao, Saurabh Tiwary, Rangan Majumder, and Li~Deng. 2016.
\newblock \href {http://arxiv.org/abs/1611.09268} {{MS} {MARCO:} {A} human generated machine reading comprehension dataset}.
\newblock \emph{CoRR}, abs/1611.09268.

\bibitem[{Rabinovich et~al.(2017)Rabinovich, Patel, Mirkin, Specia, and Wintner}]{rabinovich-etal-2017-personalized}
Ella Rabinovich, Raj~Nath Patel, Shachar Mirkin, Lucia Specia, and Shuly Wintner. 2017.
\newblock \href {https://aclanthology.org/E17-1101} {Personalized machine translation: Preserving original author traits}.
\newblock In \emph{Proceedings of the 15th Conference of the {E}uropean Chapter of the Association for Computational Linguistics: Volume 1, Long Papers}, pages 1074--1084, Valencia, Spain. Association for Computational Linguistics.

\bibitem[{Rahimi et~al.(2015)Rahimi, Shakery, and King}]{article}
Razieh Rahimi, Azadeh Shakery, and Irwin King. 2015.
\newblock \href {https://doi.org/10.1007/s10791-015-9255-1} {Multilingual information retrieval in the language modeling framework}.
\newblock \emph{Information Retrieval Journal}, 18:246--281.

\bibitem[{Santhanam et~al.(2022)Santhanam, Khattab, Potts, and Zaharia}]{10.1145/3511808.3557325}
Keshav Santhanam, Omar Khattab, Christopher Potts, and Matei Zaharia. 2022.
\newblock \href {https://doi.org/10.1145/3511808.3557325} {{PLAID}: An efficient engine for late interaction retrieval}.
\newblock In \emph{Proceedings of the 31st ACM International Conference on Information \& Knowledge Management}, CIKM '22, page 1747–1756, New York, NY, USA. Association for Computing Machinery.

\bibitem[{Tiedemann and Thottingal(2020)}]{tiedemann-thottingal-2020-opus}
J{\"o}rg Tiedemann and Santhosh Thottingal. 2020.
\newblock \href {https://aclanthology.org/2020.eamt-1.61} {{OPUS}-{MT} {--} building open translation services for the world}.
\newblock In \emph{Proceedings of the 22nd Annual Conference of the European Association for Machine Translation}, pages 479--480, Lisboa, Portugal. European Association for Machine Translation.

\bibitem[{Vanmassenhove and Hardmeier(2018)}]{vanmassenhove-hardmeier-2018-europarl}
Eva Vanmassenhove and Christian Hardmeier. 2018.
\newblock \href {https://aclanthology.org/2018.eamt-main.59} {Europarl datasets with demographic speaker information}.
\newblock In \emph{Proceedings of the 21st Annual Conference of the European Association for Machine Translation}, page 391, Alicante, Spain.

\bibitem[{Vrandečić and Krötzsch(2014)}]{42240}
Denny Vrandečić and Markus Krötzsch. 2014.
\newblock \href {http://cacm.acm.org/magazines/2014/10/178785-wikidata/fulltext} {Wikidata: A free collaborative knowledge base}.
\newblock \emph{Communications of the ACM}, 57:78--85.

\bibitem[{Wojtasik et~al.(2023)Wojtasik, Shishkin, Wołowiec, Janz, and Piasecki}]{wojtasik2023beirpl}
Konrad Wojtasik, Vadim Shishkin, Kacper Wołowiec, Arkadiusz Janz, and Maciej Piasecki. 2023.
\newblock \href {http://arxiv.org/abs/2305.19840} {{BEIR-PL}: Zero shot information retrieval benchmark for the {Polish} language}.
\newblock arXiv cs.IR 2305.19840.

\bibitem[{Yang et~al.(2017)Yang, Fang, and Lin}]{yang+:2017}
Peilin Yang, Hui Fang, and Jimmy Lin. 2017.
\newblock \href {https://doi.org/10.1145/3077136.3080721} {Anserini: Enabling the use of lucene for information retrieval research}.
\newblock In \emph{Proceedings of the 40th International ACM SIGIR Conference on Research and Development in Information Retrieval}, pages 1253--1256.

\bibitem[{Zhang et~al.(2021)Zhang, Ma, Shi, and Lin}]{zhang2021mr}
Xinyu Zhang, Xueguang Ma, Peng Shi, and Jimmy Lin. 2021.
\newblock \href {http://arxiv.org/abs/2108.08787} {{Mr.\ TyDi}: A multi-lingual benchmark for dense retrieval}.
\newblock arXiv cs.CL 2108.08787.

\end{thebibliography}
\bibliographystyle{acl_natbib}

\appendix
\onecolumn

\section{Appendix}

\begin{table}[h!]
             \centering
             \rotatebox{90}{%
                \begin{varwidth}{0.95\textheight}
                  \centering
\footnotesize 

\small 
\centering
\footnotesize 

\small 
\rotatebox{360}{ 
\begin{tabular}{m{0.7cm} *{30}{c@{\,\,\,}}c }
\toprule
\textbf & \multicolumn{6}{c}{\textbf{GERMANIC}} &&  \multicolumn{5}{c}{\textbf{ROMANCE}} && \multicolumn{6}{c}{\textbf{SLAVIC}} && \multicolumn{3}{c}{\textbf{URALIC}} && \multicolumn{2}{c}{\textbf{BALTIC}} & {\textbf{HELLENTICH}} & {\textbf{CELTC}} \\

\cmidrule{2-7}
\cmidrule{9-13}
\cmidrule{15-20}
\cmidrule{22-24}
\cmidrule{26-27}

\textbf & \textbf{EN} & \textbf{DA} & \textbf{DE} & \textbf{NL} & \textbf{SV} & \textbf{MT} && 
\textbf{RO} & \textbf{ES} & \textbf{FR} & \textbf{IT} & \textbf{PT} && 
\textbf{PL} & \textbf{BG} & \textbf{CS} & \textbf{SK} & \textbf{SL} & \textbf{HR}&& 
\textbf{HU} & \textbf{FI} & \textbf{ET} &&
\textbf{LV} & \textbf{LT} &
\textbf{EL} & 
\textbf{GA}\\

\midrule
\multicolumn{21}{l}{\textbf{One-vs-one} (Queries in one language against documents in the same language, test on the whole set.)} \\
\midrule

num$_q$ & 839 & 208 & 840 & 458 & 330 & 138 && 434 & 680 & 765 & 659 & 557 && 628 & 273 & 259 & 236 & 205 & 311 && 404 & 251 & 52 && 75 & 99 & 360 & 14 \\

num$_d$ & 7123 & 292 & 3433 & 897 & 531 & 178 && 794 & 2371 & 2779 & 1829 & 1176 && 1841 & 408 & 397 & 348 & 270 & 524 && 614 & 405 & 58 && 89 & 115 & 707 & 16 \\

num$_r$ & 7123 & 292 & 3433 & 897 & 531 & 178 && 794 & 2371 & 2779 & 1829 & 1176 && 1841 & 408 & 397 & 348 & 270 & 524 && 614 & 405 & 58 && 89 & 115 & 707 & 16 \\


\textbf{MRR} & \textbf{73.79} & \textbf{45.51} & \textbf{56.70} & \textbf{39.02} & \textbf{45.77} & \textbf{65.17} && \textbf{42.58} & \textbf{54.25} & \textbf{56.51} & \textbf{47.68} & \textbf{56.40} && \textbf{47.12} & \textbf{44.70} & \textbf{47.12} & \textbf{44.51} & \textbf{45.47} & \textbf{39.83} && \textbf{39.99} & \textbf{46.46} & \textbf{41.59} && \textbf{58.04} & \textbf{50.03} & \textbf{39.58} & \textbf{72.22} \\

\midrule
\multicolumn{21}{l}{\textbf{One-vs-many} (Queries in one language against documents in many languages, test on semantically consistent topics.)} \\
\midrule

num$_q$ & 100 & 100 & 100 & 100 & 100 & 100 && 100 & 100 & 100 & 100 & 100 && 100 & 100 & 100 & 100 & 100 & 100 && 100 & 100 & 100 && 100 & 100 & 100 & 100\\
num$_d$ & 22K & 22K & 22K & 22K & 22K & 22K && 22K & 22K & 22K & 22K & 22K && 22K & 22K & 22K & 22K & 22K & 22K && 22K & 22K & 22K && 22K & 22K & 22K & 22K\\

num$_r$ & 2902 & 1997 & 1996 & 1996 & 1996 & 1996 && 1996 & 1924 & 1996 & 1997 & 1996 && 1996 & 1996 & 1997 & 1996 & 1996 & 1923 && 1996 & 1997 & 1909 && 1996 & 1997 & 1996 & 1992\\

\textbf{MRR} & \textbf{62.79} & \textbf{16.15} & \textbf{28.27} & \textbf{20.88} & \textbf{19.40} & \textbf{0.40} && \textbf{16.10} & \textbf{22.57} & \textbf{24.22} & \textbf{14.24} & \textbf{18.70} &&\textbf{4.80} & \textbf{7.57} & \textbf{9.52} & \textbf{12.35} & \textbf{6.46} & \textbf{10.41} && \textbf{7.51} & \textbf{17.61} & \textbf{7.31} && \textbf{7.95} & \textbf{8.94} & \textbf{11.16} & \textbf{5.21}\\


\bottomrule
\end{tabular}}
\caption{Details of Multi-EuP for the all 24 spoken EU official languages, in terms of the number of queries ($q$), documents ($d$) and relevance judgements ($r$). Results are for BM25 in one-vs-one and one-vs-many settings based on MRR$@$100 (\%). Note that as each document has a unique topic which in turn defines the relevance judgements, num$_d$= num$_r$ in the one-vs-one setting.}

\label{AP:Multi-EuP-stats}

\end{varwidth}}
\end{table}


\end{document}